%% file: Main.tex
\documentclass{article}
\usepackage{spconf,amsmath,graphicx}
\usepackage{algorithm}
\usepackage{algorithmic}
\usepackage{amsfonts}
\usepackage{amsmath}
\usepackage{xcolor}
\usepackage{booktabs}
\usepackage{pifont}
\usepackage{makecell}
\usepackage{array}

\usepackage{hyperref}

\title{FC-VFI: FAITHFUL AND CONSISTENT VIDEO FRAME INTERPOLATION \\ FOR HIGH-FPS SLOW MOTION VIDEO GENERATION}

\name{Ganggui Ding$^{\star}$ \qquad Hao Chen$^{\star} \thanks{HC and XX are co-corresponding authors.}$ \qquad Xiaogang Xu $^{\dagger} \thanks{This work was supported by the ``Pioneer'' and ``Leading Goose'' R\&D Program of Zhejiang (Grant No. 2025C01011) and the National Natural Science Foundation of China (No. 62576315).} $}

\address{$^{\star}$ Zhejiang University $^{\dagger}$ Chinese University of Hong Kong} 

\begin{document}
\ninept
\maketitle
\begin{abstract}
Large pre-trained video diffusion models excel in video frame interpolation but struggle to generate high fidelity frames due to reliance on intrinsic generative priors, limiting detail preservation from start and end frames. Existing methods often depend on motion control for temporal consistency, yet dense optical flow is error-prone, and sparse points lack structural context. In this paper, we propose FC-VFI for faithful and consistent video frame interpolation, supporting 4× and 8× interpolation, boosting frame rates from 30 FPS to 120 and 240 FPS at 2560 × 1440 resolution while preserving visual fidelity and motion consistency. We introduce a temporal modeling strategy on the latent sequences to inherit fidelity cues from start and end frames and leverage semantic matching lines for structure-aware motion guidance, improving motion consistency. Furthermore, we propose a temporal difference loss to mitigate temporal inconsistencies. Extensive experiments show FC-VFI achieves high performance and structural integrity across diverse scenarios.
\end{abstract}
\begin{keywords}
Computer Vision, Video Frame Interpolation, Video Generation, Diffusion Model, Generative Model
\end{keywords}

\begin{figure*}[t]
  \centering
  \includegraphics[width=2.0\columnwidth]{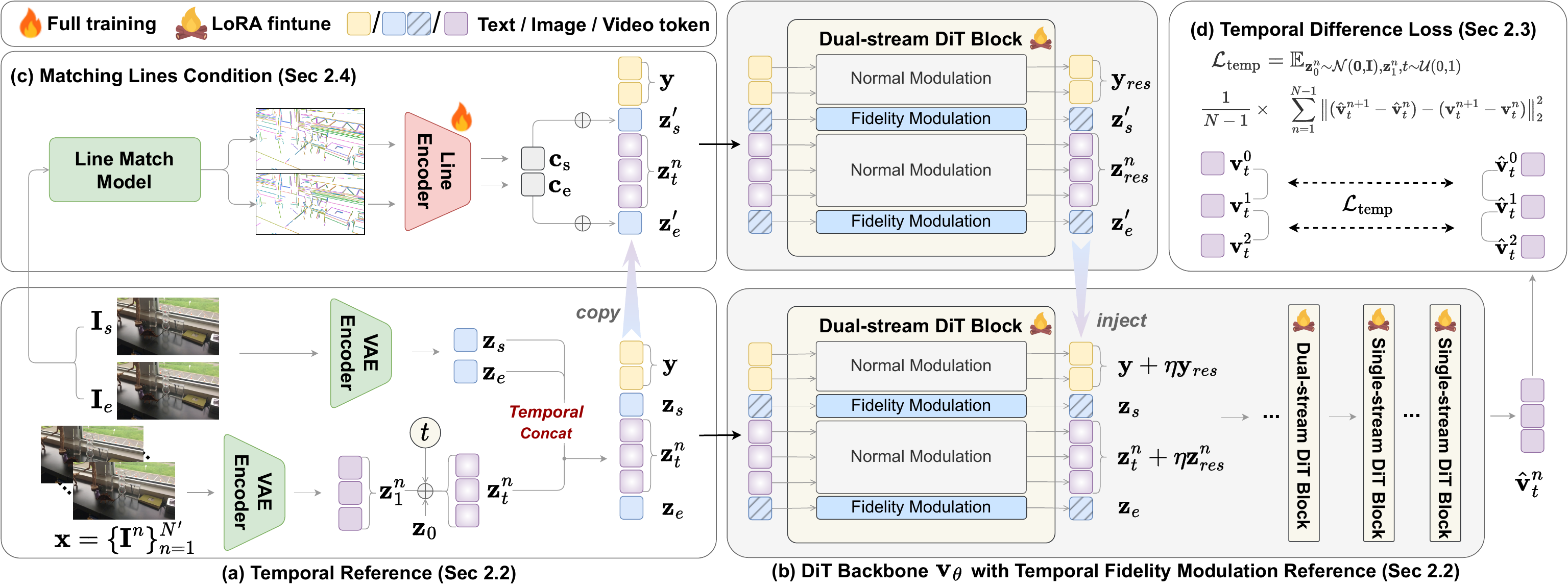} 
  \vspace{-0.1in}
  \caption{ \textbf{Overview of FC-VFI's training pipeline.}
    (a) We model temporal references by concatenating the noisy latent \( \mathbf{z}_t^n \) with the start and end image latents \( \mathbf{z}_s \) and \( \mathbf{z}_e \) along the temporal dimension, enabling the denoising process to reference both boundaries.
    (b) We apply \textit{fidelity modulation} by performing a timestep-dependent modulation \( t^* \) into \( \mathbf{z}_s \) and \( \mathbf{z}_e \), which enhances reference stability.
    (c) Semantic matching lines featrues \( \mathbf{c}_s \) and \( \mathbf{c}_e \) are extracted and encoded from the start frame \( \mathbf{I}_s \) and end frame \( \mathbf{I}_e \), and are element-wise added to \( \mathbf{z}_s \) and \( \mathbf{z}_e \), resulting in enhanced latents \( \mathbf{z}_s' \) and \( \mathbf{z}_e' \). These are then processed via a copied DiT block to produce \( \mathbf{z}_{\text{res}}^n \), which is injected back into the main backbone.
    (d) The prediction \( \hat{\mathbf{v}}_t^n \) is supervised with a temporal difference loss \( \mathcal{L}_{\text{temp}} \).}
  \label{fig_pipeline_overview}
  \vspace{-0.2in}
\end{figure*}

\section{Introduction}
\label{sec:intro}

Video Frame Interpolation (VFI)~\cite{seo2025bim} is a fundamental task in computer vision that aims to synthesize intermediate frames between given start and end frames. It has broad applications, including animation production~\cite{xu2023conditional} and slow motion video generation~\cite{huang2022real}.

\textbf{Background.} Traditional VFI methods~\cite{huang2022real,seo2025bim} typically rely on motion representations to synthesize intermediate frames, e.g., estimating optical flow between the start and end frames. However, these approaches often struggle in complex scenes where dense motion is difficult to estimate accurately. To address this limitation, diffusion-based methods have recently emerged~\cite{wang2024framer,zhu2025generative}, leveraging generative capabilities to handle such challenging cases. These approaches generally encode the start and end frames into a latent space and use them as conditioning inputs to predict the latent representation of the intermediate frames. 

\begin{figure*}[t]
  \centering
\includegraphics[width=\textwidth]{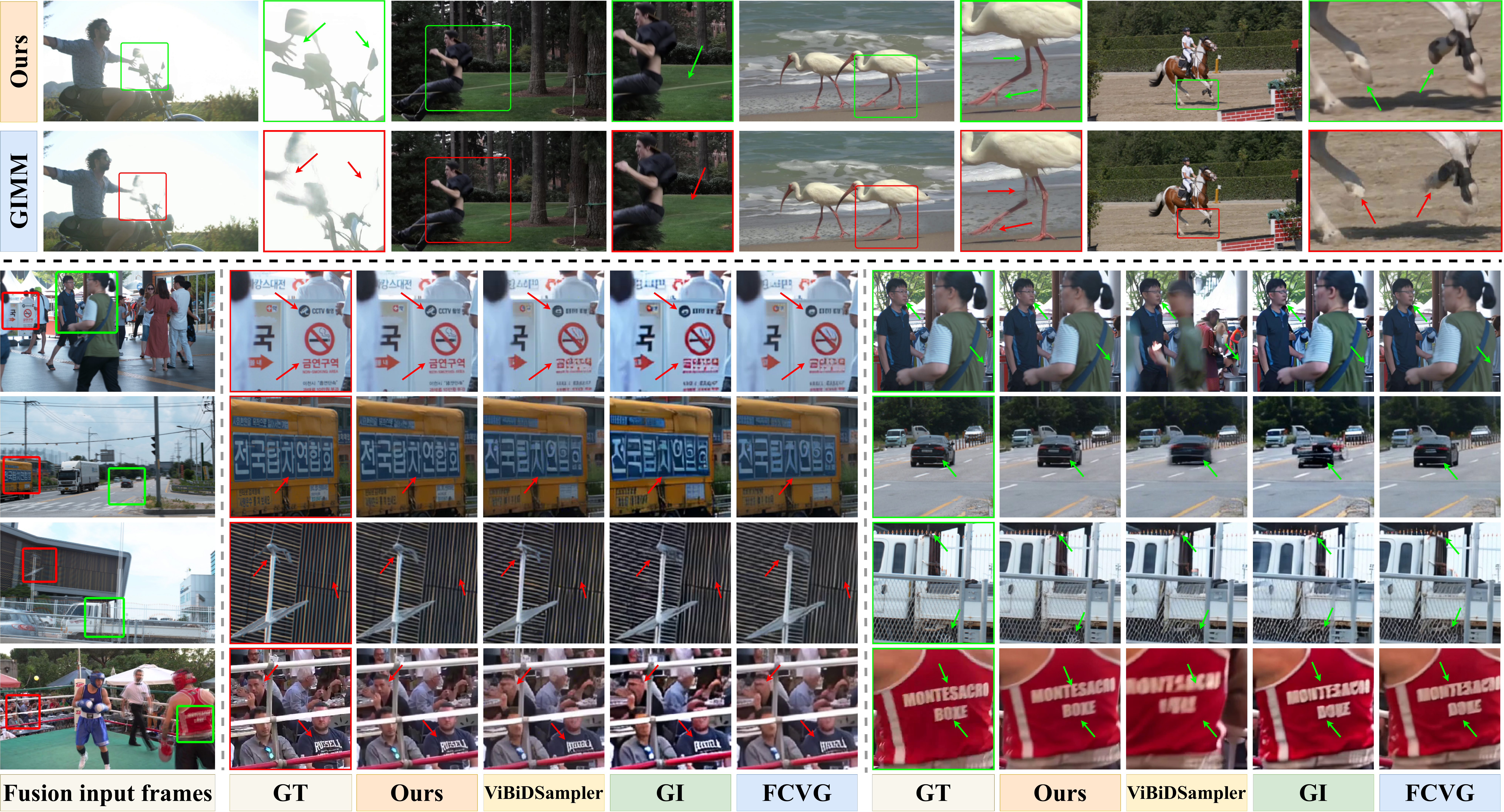} 
\vspace{-0.2in}
\caption{
    \textbf{Qualitative comparison of interpolation results.}
      (Top) Comparison with GIMM-VFI~\cite{guo2024generalizable} on DAVIS-2017 \cite{pont20172017} at \(2560 \times 1440\) resolution under \(8\times\) interpolation. Ours better handles challenging conditions such as high-contrast lighting, small objects, and occlusion, avoiding artifacts like ghosting and structural distortion.
      (Bottom) Comparison with diffusion-based methods (GI~\cite{wang2024generative}, ViBiDSampler~\cite{yang2024vibidsampler}, FCVG~\cite{zhu2025generative}) on X-Test~\cite{sim2021xvfi} and DAVIS-2017 at \(1024 \times 576\) resolution under \(8\times\) interpolation. FC-VFI preserves finer details (e.g., text, license plates, building textures), while other methods suffer from motion ambiguity and temporal artifacts.
    }
  \label{fig:qualitative_comparison}
\vspace{-0.2in}
\centering
\end{figure*}

\textbf{The challenges of current diffusion-based VFI methods.} However, current diffusion-based VFI methods still exhibit notable weaknesses, e.g., the fidelity issue caused by generative models~\cite{pan2025boosting,zhang2025motion} and the temporal inconsistency of interpolated results~\cite{danier2024ldmvfi}. Fidelity issues often manifest as artifacts or structural distortions in the intermediate frames. For instance, a car may appear deformed compared to its shape in the start and end frames, leading to flickering and perceptual inconsistency, as shown in Fig.~\ref{fig:qualitative_comparison}. 

Moreover, despite the strong generative priors of diffusion models, the motion in interpolated sequences may still exhibit inaccuracies.
To mitigate these issues, some approaches incorporate optical flow or sparse correspondence points to guide motion~\cite{zhang2025motion,wang2024framer}. However, optical flow estimation can be error-prone in complex scenes~\cite{xu2022mtformer, zhang2025motion}, potentially degrading interpolation quality, while sparse points are insufficient to capture detailed object structure.

Furthermore, the efficiency of current diffusion-based VFI methods remains a concern. Many approaches generate video frames from the start and end frames independently, and subsequently merging them to produce the final result through bidirectional time-reversal fusion~\cite{feng2024explorative,yang2024vibidsampler,wang2024generative}. Some methods even require additional re-noising steps~\cite{feng2024explorative,yang2023object}, further increasing computational overhead.

\textbf{Our model.}
In this paper, we propose FC-VFI for \underline{f}aithful and \underline{c}onsistent \underline{v}ideo \underline{f}rame \underline{i}nterpolation. Finetuned from a pre-trained large-scale I2V model, our method supports 4× and 8× interpolation at resolutions up to 2560 × 1440, eliminates the need for bidirectional inference, compared to previous diffusion-based methods.

To improve visual fidelity, we propose Temporal Fidelity Modulation Reference (TFMR), a novel temporal modeling strategy. Unlike existing diffusion-based methods that typically perceive conditional latents via channel-wise concatenation~\cite{blattmann2023stable,ding2024freecustom, xu2022hierarchical}, TFMR combines the noise of the intermediate frames with the start and end frames along temporal dimension, and perform fidelity modulation on both boundary frames. This design ensures that the intermediate frames consistently reference features from both endpoints throughout the generation process. 

To mitigate the near-static behavior between interpolated adjacent frames, we introduce a temporal difference loss that explicitly aligns the predicted motion difference between consecutive frames with that of the ground truth.

To enhance temporal consistency, we introduce a novel conditioning mechanism based on semantic matching lines, which offers greater robustness than optical flow by focusing only on key motion boundaries, and provide richer structural information than sparse points by describing the ojbect shape. 

In summary, our contributions are threefold:
\begin{itemize}
\item We propose an effective training strategy to fine-tune pre-trained I2V diffusion models into VFI networks, enabling practical interpolation, e.g., from 30 FPS to 120 FPS and 240 FPS for 2560 × 1440 videos.
\item We introduce a novel temporal modeling strategy to address fidelity issues in interpolation, alongside a matching lines condition control mechanism and a temporal difference loss to achieve consistent and accurate motion.
\item Extensive qualitative and quantitative experiments demonstrate the superiority of our model for 4× and 8× interpolation tasks, highlighting its robust stability and structural consistency, as shown in Fig.~\ref{fig:qualitative_comparison} and Table~\ref{tab_quantitative_results}.
\end{itemize}

\section{Method}

\subsection{Preliminary: Flow Matching Model}
Flow Matching (FM)~\cite{lipman2022flow} is a generative framework that learns a continuous mapping between Gaussian noise and the target data distribution. Given a video $\mathbf{x} \in \mathbb{R}^{N' \times 3 \times H' \times W'}$, it is first encoded by a VAE~\cite{kingma2013auto} encoder $\mathcal{E}$ into a latent $\mathbf{z}_1 \in \mathbb{R}^{N \times C \times H \times W}$, where $C$ is channel dimension, $N = N'/\lambda_t$, $H = H'/\lambda_s$, and $W = W'/\lambda_s$ ($\lambda_t, \lambda_s$ as temporal and spatial compression ratios of the $\mathcal{E}$, respectively). A noisy latent is then obtained as:
$
\mathbf{z}_t = (1 - t)\mathbf{z}_1 + t \mathbf{z}_0, \quad \mathbf{z}_0 \sim \mathcal{N}(\mathbf{0}, \mathbf{I}),
$
where $t \in [0,1]$. The training objective for a I2V model~\cite{kong2024hunyuanvideo,blattmann2023stable} minimizes the discrepancy between the predicted velocity $\mathbf{v}_\theta$ and the true velocity $\mathbf{v}_t = \mathbf{z}_0 - \mathbf{z}_1$:
\begin{equation}
\mathcal{L}_{\text{flow}} = \mathbb{E}_{\mathbf{z}_0, \mathbf{z}_1, t \sim \mathcal{U}(0,1)} \left\| \mathbf{v}_t - \mathbf{v}_\theta(\mathbf{z}_t, \mathbf{z}_s, y, t) \right\|_2^2,
\end{equation}
where $\mathbf{z}_s$ is the latent representation of the start frame, $y$ is the corresponding text prompt for the I2V generation task.

\subsection{Temporal Fidelity Modulation Reference}
\label{sec:2}

The VFI task generates \(N'\) intermediate frames \(\mathbf{x}=\{\mathbf{I}^n\}_{n=1}^{N'}\) between the start frame \(\mathbf{I}_\text{s}\) and end frame \(\mathbf{I}_\text{e}\). We propose a temporal reference strategy by concatenating the latents of boundary frames \(\mathbf{z}_s=\mathcal{E}(\mathbf{I}_\text{s})\), \(\mathbf{z}_e=\mathcal{E}(\mathbf{I}_\text{e})\), and noisy intermediate latents \(\{\mathbf{z}^n\}_{n=1}^N\) along the temporal dimension, forming \(\mathcal{Z}=\{\mathbf{z}_s,\mathbf{z}^n,\mathbf{z}_e\}\). Intermediate frames are generated by denoising \(\mathbf{z}^n\) guided by \(\mathbf{z}_s,\mathbf{z}_e\).

In DiT~\cite{peebles2023scalable}, timestep \(t\) and text condition \(y\) jointly control modulation parameters scale, shift, and gate \((\gamma,\beta,\alpha)\), with \(t\) controlling the denoising intensity: stronger at the start (\(t=1\)) and weaker at the end (\(t=0\)) . However, applying uniform velocity prediction \(\mathbf{v}_\theta(\mathcal{Z},y,t)\) to all latents in \(\mathcal{Z}\) may perturb clean boundary latents.

\textbf{Fidelity Modulation.}
To preserve the integrity of the boundary frames, we assign a fixed timestep \(t^* = 0\) for \(\mathbf{z}_s\) and \(\mathbf{z}_e\), corresponding to the noise-free state, while intermediate frames retain the standard schedule. Then, the predicted velocity is:
\begin{equation}
\mathbf{v}_\theta(\mathcal{Z}(t^*),y,t)=\mathbf{v}_\theta(\{\mathbf{z}_s(t^*),\mathbf{z}^n_t,\mathbf{z}_e(t^*)\},y,t).
\end{equation}
This preserves boundary fidelity and constrains intermediate motion. Thus, the flow matching loss is redefined over intermediate frames:
\begin{equation}
\mathcal{L}_{\text{flow}}=\mathbb{E}_{\mathbf{z}^n_0,\mathbf{z}^n_1,t}\sum_{n=1}^N \left\|\mathbf{v}^n_t-\mathbf{v}_\theta(\mathcal{Z}(t^*),y,t)\right\|_2^2.
\end{equation}

\subsection{Temporal Difference Loss}
\label{sec:3}

Small motion amplitudes often yield near-static interpolations. To mitigate this, we introduce a temporal difference loss encouraging dynamic distinctions among consecutive latents:
$$
\mathcal{L}_{\text{temp}}=\mathbb{E}_{\mathbf{z}^n_0,\mathbf{z}^n_1,t}\frac{1}{N-1}\sum_{n=1}^{N-1}\left\|(\hat{\mathbf{v}}^{n+1}_t-\hat{\mathbf{v}}^n_t)-(\mathbf{v}^{n+1}_t-\mathbf{v}^n_t)\right\|_2^2,
$$
where $\hat{\mathbf{v}}$ is the predicted velocity by \(\mathbf{v_\theta}\). This alleviates the near-static behavior between adjacent frames and promotes smoother motion transitions. Consequently, the final training objective is:
\begin{equation}
\mathcal{L}=\mathcal{L}_{\text{flow}}+\omega\mathcal{L}_{\text{temp}}.
\label{eq_loss}
\end{equation}

\subsection{Matching Lines Condition}
\label{sec:4}
Maintaining structural stability under large camera motion or rapid object movement is challenging. Existing methods use dense optical flow~\cite{zhang2025motion} or sparse point trajectories~\cite{wang2024framer} to extract semantic information, but often fail to capture accurate object structures. FCVG~\cite{zhu2025generative} introduces frame-wise conditions via ControlNeXt~\cite{peng2024controlnext} architecture, while it may incompatible with modern large-scale I2V models~\cite{kong2024hunyuanvideo}, which employ temporally compressed VAEs (\(\lambda_t>1\)), where a single latent frame influences multiple image frames during decoding. Thus, directly applying frame-wise features to video latent risks introducing incorrect structural information.

To overcome this, we extract semantically consistent line pairs from the start frame \(\mathbf{I}_\text{s}\) and end frame \(\mathbf{I}_\text{e}\) using GlueStick~\cite{pautrat2023gluestick}. A lightweight ResNet-based line encoder encodes them into condition features \(\mathbf{c}_s\) and \(\mathbf{c}_e\), which are fused with boundary latents through element-wise addition: \(\mathbf{z}_s'=\mathbf{z}_s+\mathbf{c}_s\), \(\mathbf{z}_e'=\mathbf{z}_e+\mathbf{c}_e.\)
Then, the updated sequence \(\mathcal{Z}'=\{\mathbf{z}_s',\mathbf{z}^n,\mathbf{z}_e'\}\) is processed by a single replicated DiT block, yielding residual features \(\mathbf{z}^n_{\text{res}}\). After normalization, they are injected into the backbone by:
\begin{equation}
\mathbf{z}^n_{\text{updated}}=\mathbf{z}^n+\eta \mathbf{z}^n_{\text{res}},
\label{z_res_add}
\end{equation}
where \(\eta\) controls injection strength. This strategy avoids using single-frame-scale features to control multi-frame-scale video latents, preventing structural interference and enhancing object structural stability in fast motion scenarios. We introduce only 2.7\% extra parameters—significantly lighter than ControlNet~\cite{zhang2023adding} architecture, which duplicates a significant portion of the backbone modules.

\input{tables/quantitative_results.tex}
\input{tables/ablation.tex}

\input{tables/computation_efficiency.tex}

\section{Experiments}
\subsection{Experimental Setup}

\noindent\textbf{Training Datasets.}
We construct a diverse mixed dataset from REDS~\cite{nah2019ntire} and Adobe240~\cite{shen2020blurry}, covering a wide range of real-world scenarios captured by various devices. All videos are resized to $1280 \times 720$ and uniformly segmented into non-overlapping 9-frame clips, resulting in 13,200 clips at 120 FPS and 13,653 clips at 240 FPS, used for training 4× and 8× interpolation, respectively.

\textbf{Evaluation Metrics.}
To comprehensively evaluate our model's VFI performance, we adopt FVD~\cite{unterthiner2018towards} for video quality and motion coherence, and use FID~\cite{heusel2017gans}, LPIPS~\cite{zhang2018unreasonable}, PSNR, and SSIM~\cite{wang2004image} to evaluate the visual fidelity and perceptual quality of the interpolated frame individually, following FCVG \cite{zhu2025generative} and Framer \cite{wang2024framer}. Our test dataset comprises 97 high-frame-rate videos from X-Test~\cite{sim2021xvfi}, BVI-DVC~\cite{ma2021bvi}, and DAVIS-2017~\cite{pont20172017}, yielding 326 start-and-end frame input pairs for comprehensive evaluation.

\textbf{Implementation Details.}
Our model is based on the FM-based HunyuanVideo-I2V model, fine-tuned using LoRA~\cite{hu2022lora} with a rank of 64. 
We set the hyperparameters in Eq.~\ref{eq_loss} and Eq.~\ref{z_res_add} to \(\omega=0.8\) and \(\eta=1.0\).
The depth of the replicated DiT block is 1, adding only 2.7\% parameters. Training is performed in two stages: (1) fine-tuning the DiT backbone with LoRA for 1 epoch, and (2) full-training the line encoder and fine-tuning image-attention-related modules in DiT blocks with LoRA for 0.5 epochs.

\subsection{Comparisons with SOTA Models}
To evaluate the performance of our proposed FC-VFI model, we conduct a comparative analysis with SOTA video frame interpolation methods developed in the past two years, encompassing both optical-flow-based (GIMM-VFI \cite{guo2024generalizable}) and diffusion-based approaches (FCVG \cite{zhu2025generative}, ViBiDSampler \cite{yang2024vibidsampler}, and GI \cite{wang2024generative}). 

\textbf{Quantitative Results.}
As shown in Table~\ref{tab_quantitative_results}, we conduct quantitative comparisons across two resolution settings: 2560 × 1440 and $1024\times576$. Under the high-resolution setting, our method achieves competitive performance compared to GIMM-VFI.
In contrast, when evaluated under $1024\times576$ resolution, our method clearly surpasses all recent diffusion-based approaches across all five metrics, under both 4× and 8× interpolation settings, reflecting superior reconstruction fidelity and perceptual quality. 

\textbf{Qualitative Results.}
As illustrated in Fig.~\ref{fig:qualitative_comparison} (Top), our method exhibits strong visual performance compared to GIMM-VFI. Notably, although our model is trained on \(1280 \times 720\) resolution videos, it generalizes well to high-resolution scenarios, since it employs TFMR to propagates fidelity information from the boundary frames to intermediate frames.
Fig.~\ref{fig:qualitative_comparison} (Bottom) presents visual comparisons with recent diffusion-based methods, FC-VFI better reconstructs fine-grained details such as billboard text, license plates, and dense architectural patterns. 
These results validate the advantage of our method in preserving both motion consistency and visual fidelity. 

\subsection{Computation Efficiency}

Although FC-VFI is built upon the large-scale HunyuanVideo-I2V backbone with 13B parameters, it achieves efficient inference by requiring only a few denoising steps. In particular, the proposed TFMR reduces the denoising complexity at each timestep, enabling high-quality interpolation with only 10 steps. As shown in Table~\ref{tab_computation_efficiency}, we compare with recent diffusion-based methods (GI, ViBiDSampler, and FCVG) at \(1024 \times 576\). These methods typically rely on multi-stage denoising process or high numbers of function evaluations (NFE). In contrast, FC-VFI achieves faster inference with significantly fewer steps, even at a higher resolution of \(1280 \times 720\).

\begin{figure}[t]
    \centering
    \includegraphics[width=\linewidth]{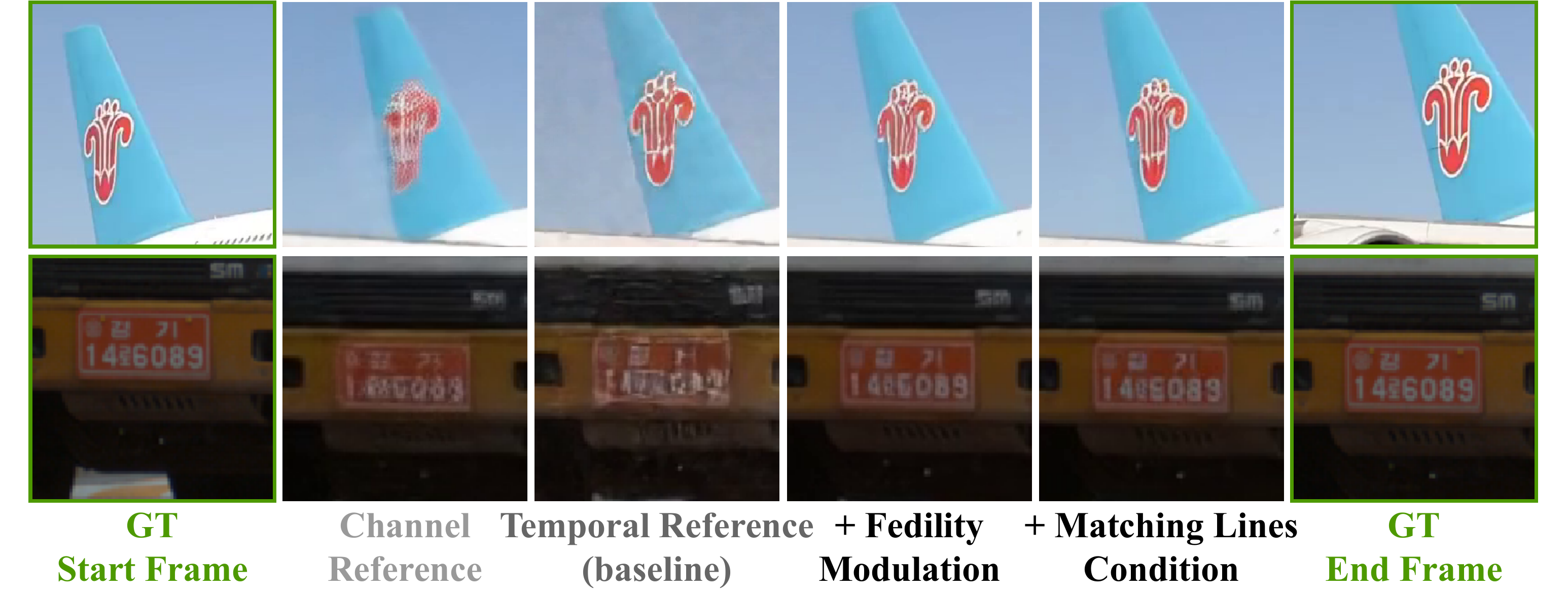}
    \vspace{-0.2in}
    \caption{Ablation results of our method visualized under different module configurations. The displayed intermediate frame is closer to the end frame.}
    \label{fig_ablation}
     \vspace{-0.1in}
\end{figure}

\subsection{Ablation Study}

We conduct ablations to assess the contribution of each component, as shown in Table~\ref{tab_ablation} and Fig.~\ref{fig_ablation}. All experiments are performed on testset at \(1280 \times 720\) resolution. We first compare with a naive channel reference that feeds start and end frame through channel concatenation. This strategy degrades structural consistency and visual detail (third vs. fourth row in Table~\ref{tab_ablation}). In contrast, the temporal reference baseline concatenates start and end frame along temporal dimension, providing clearly motion boundary.

Building on temporal reference, the fidelity modulation mechanism further improves structural fidelity by explicitly correcting timestep noise in boundary frame conditioning (comparing the fourth and fifth row in Table~\ref{tab_ablation}).
Adding the temporal difference loss \(\mathcal{L}_{\text{temp}}\) leads to smoother transitions across intermediate frames, enhancing temporal coherence (fifth row v.s. sixth row in Table~\ref{tab_ablation}). Finally, incorporating the matching lines condition yields the most refined results by injecting line-level correspondence priors (sixth row v.s. seventh row in Table~\ref{tab_ablation}), improving detail restoration such as edges and textures.

\section{Conclusion}

In this paper, we propose FC-VFI, a diffusion-based video frame interpolation (VFI) framework finetuned from a pre-trained IV2 model (e.g., HunyuanVideo-I2V). Our framework introduces Temporal Fidelity Modulation Reference (TFMR), which propagates the fidelity information from the start and end frames using temporal concatenation and fidelity modulation, enabling efficient inference within only 10 denoising steps. In addition, we design a novel temporal difference loss and a matching-lines-based control strategy to further enhance temporal consistency and reduce artifacts. Extensive experiments on public datasets demonstrate the effectiveness of our method, particularly in high-FPS video frame interpolation.

\bibliographystyle{IEEEbib}
\bibliography{refs} 


\clearpage
\onecolumn
\appendix
\section{Supplementary Materials}

\subsection{Additional Qualitative Results}
\label{sec:appendix_qualitative}

\begin{figure}[htbp]
    \centering
    \includegraphics[width=\textwidth]{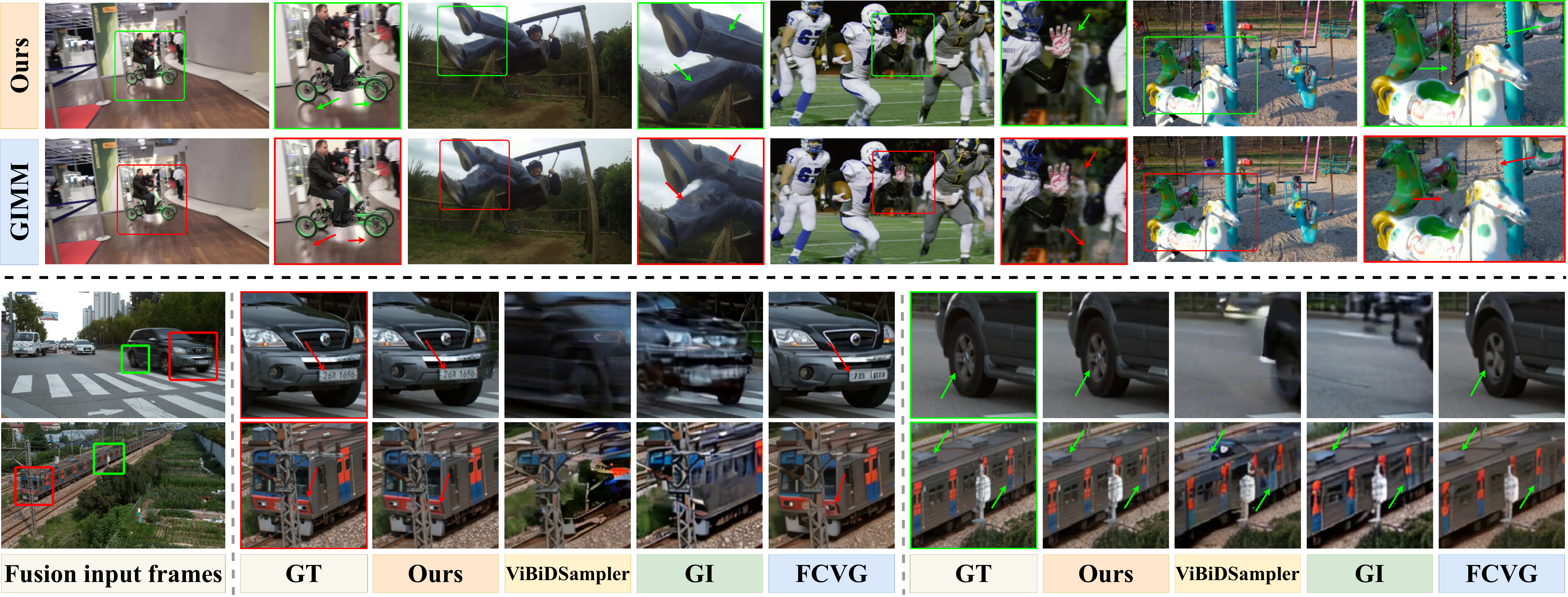} 
    \caption{
      Additional qualitative comparison of interpolation results.
      (Top) Visual comparisons with GIMM-VFI~\cite{guo2024generalizable} at $2560 \times 1440$ resolution under $8\times$ interpolation. Tested on additional challenging scenes, our FC-VFI effectively suppresses structural distortion and ghosting artifacts. 
      (Bottom) Visual comparisons with recent diffusion-based methods (GI~\cite{wang2024generative}, ViBiDSampler~\cite{yang2024vibidsampler}, and FCVG~\cite{zhu2025generative}) at $1024 \times 576$ resolution under $8\times$ interpolation. FC-VFI consistently demonstrates superior capability in recovering fine details (e.g., complex boundaries and textual patterns) and maintaining temporal consistency compared to the baselines (Sec. \ref{sec:appendix_qualitative}).
    }
    \vspace{0.2in}
    \label{fig:qualitative_comparison_appendix}
  
    \centering
    \begin{minipage}{0.48\textwidth}
        \centering
        \includegraphics[width=\linewidth]{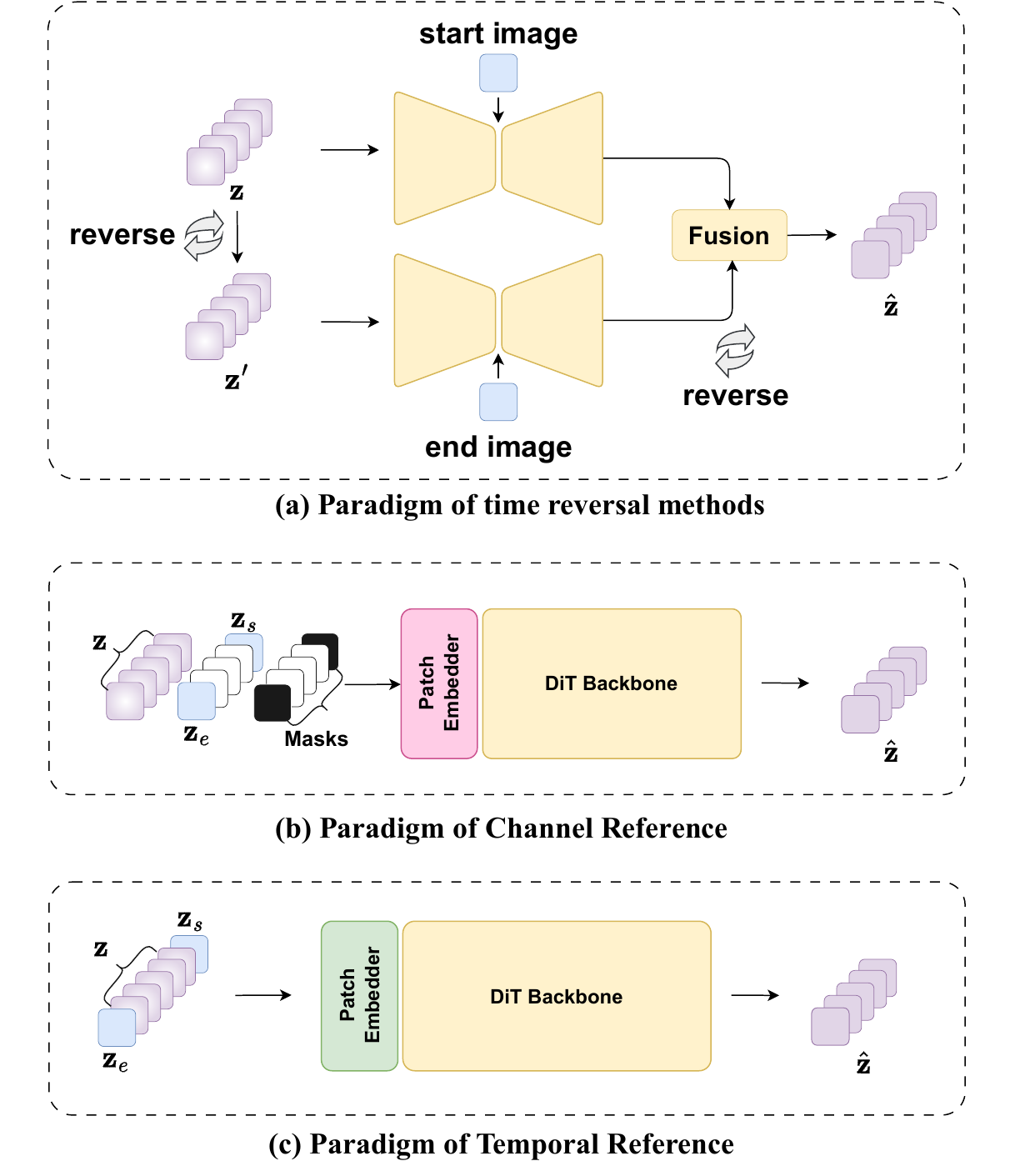}
        \caption{Comparison of Time Reversal, Channel Reference, and Temporal Reference paradigms for diffusion-based video frame interpolation (Sec. \ref{sec_difference_with_other_methods}).}
        \label{fig_difference_with_other_methods}
    \end{minipage}
    \hfill 
    \begin{minipage}{0.48\textwidth}
        \centering
        \includegraphics[width=\linewidth]{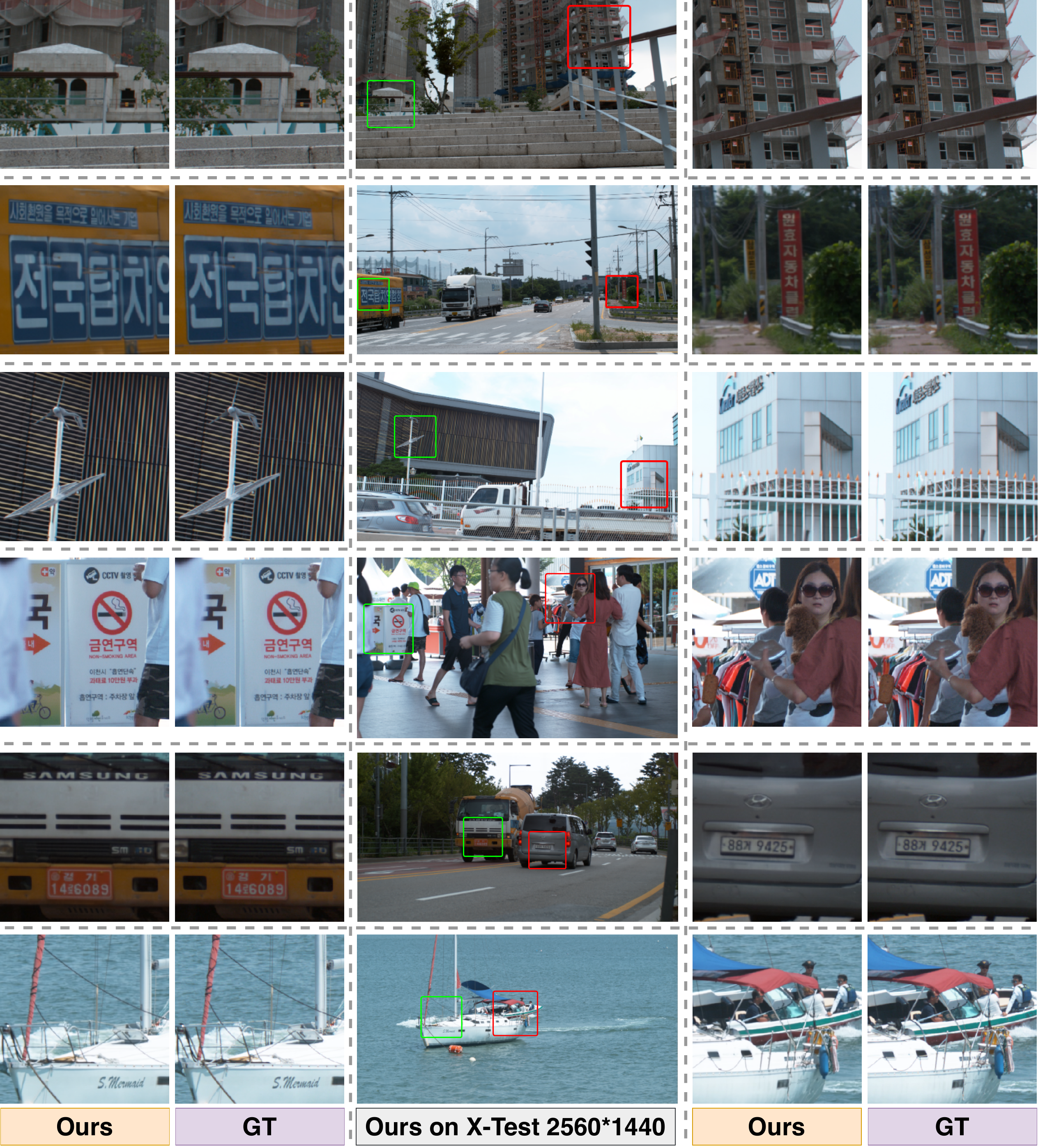}
        \caption{Qualitative results of FC-VFI on the X-Test dataset \cite{sim2021xvfi} at 2560$\times$1440 (zero-shot). The center shows a thumbnail of the interpolated frame (non-native resolution), with the left panel comparing our result (green box) to ground-truth and the right panel comparing our result (red box) to ground-truth (Sec. \ref{sec_high_res}).}
        \label{fig_2K_res}
    \end{minipage}
\end{figure}

To further demonstrate the robust stability and structural consistency of our proposed FC-VFI, we provide additional qualitative comparisons with state-of-the-art methods in Figure~\ref{fig:qualitative_comparison_appendix}. 
Following the identical evaluation setting as the main text, we compare our method against the representative optical-flow-based method GIMM-VFI~\cite{guo2024generalizable} at high resolution ($2560 \times 1440$), and recent diffusion-based methods (GI~\cite{wang2024generative}, ViBiDSampler~\cite{yang2024vibidsampler}, and FCVG~\cite{zhu2025generative}) at $1024 \times 576$ resolution. 

As illustrated in the newly added diverse test scenes, FC-VFI consistently preserves fine-grained details and structural integrity, even in highly challenging scenarios involving complex textures, extreme object motions, and severe occlusions. In contrast, the baseline approaches still suffer from noticeable visual degradation, including structural distortion, motion ambiguity, and ghosting artifacts. These supplementary visual results further validate the generalizability and superiority of our Temporal Fidelity Modulation Reference (TFMR) and matching-lines condition mechanisms.

\subsection{Paradigm Comparison with Diffusion Based Methods.}
\label{sec_difference_with_other_methods}

Previous diffusion-based video frame interpolation methods can be categorized into Time Reversal and Channel Reference paradigms, as depicted in Fig. \ref{fig_difference_with_other_methods}.

\noindent\textbf{Time Reversal methods}, including TRF \cite{feng2024explorative}, GI \cite{wang2024generative}, ViBiDSampler \cite{yang2024vibidsampler}, RE-VDM \cite{chen2025repurposing}, and FCVG \cite{zhu2025generative}, rely on conditional diffusion models, performing two image-to-video (I2V) generations conditioned on the start (\(\mathbf{z}_s\)) and end (\(\mathbf{z}_e\)) frames, respectively, followed by a fusion mechanism. These methods typically require two denoising passes, significantly increasing computational complexity. Some incorporate re-noising (TRF \cite{feng2024explorative}, GI \cite{wang2024generative}) to smooth intermediate frames, further elevating the number of function evaluations (NFE) and causing instability, particularly motion reversal artifacts in scenes with fast-moving objects or complex camera dynamics, compromising structural integrity and motion continuity. 

\noindent\textbf{Channel Reference methods} integrate boundary frame information via channel concatenation or element-wise feature addition, as seen in Framer \cite{wang2024framer}. These methods require the base I2V model to support channel-wise conditioning, limiting compatibility with models like HunyuanVideo-I2V \cite{kong2024hunyuanvideo}, which use alternative conditioning mechanisms. Adapting HunyuanVideo-I2V for channel reference necessitates training a new Patch Embedder, disrupting the pre-trained prior of its large-scale DiT backbone and degrading interpolation quality, such as visual fidelity and structural integrity. This constraint restricts deployment flexibility for such methods.

\noindent Our proposed \textbf{Temporal Reference paradigm} concatenates boundary frames \(\mathbf{z}_s, \mathbf{z}_e\) in the temporal dimension, providing explicit, high-fidelity reference conditions for the noise latent, aligning with video temporal modeling. This temporal concatenation seamlessly integrates with the token-wise input mechanism of modern DiT-based models, enhancing the denoising process's initialization. The Temporal Fidelity Modulation Reference (TFMR) mechanism enhances motion trajectory and detail capture via modulated boundary frame information, enabling robust high-resolution interpolation (e.g., 2560$\times$1440) with reduced motion blur and structural distortion in challenging scenarios like occlusions or weak textures. Compatible with diverse architectures, including HunyuanVideo-I2V \cite{kong2024hunyuanvideo}, WAN \cite{wan2025}, and SVD \cite{blattmann2023stable}, FC-VFI with TFMR offers a versatile, high-performance solution for video frame interpolation, achieving superior quality and generalization.

\subsection{High Resolution Frames Interpolation.}
\label{sec_high_res}
Our Temporal Fidelity Modulation Reference (TFMR) mechanism enables high-fidelity interpolation of intermediate frames by leveraging modulated boundary frame information. Despite pre-training and fine-tuning the base FC-VFI model on datasets with a maximum resolution of 1280$\times$720, our approach generalizes robustly to 2560$\times$1440, achieving high-quality zero-shot interpolation. Fig. \ref{fig_2K_res} presents qualitative results on the X-Test dataset \cite{sim2021xvfi} at 2560$\times$1440, demonstrating FC-VFI's superior performance in challenging scenarios, including text, dense textures, buildings, faces, and license plates. Compared to ground-truth, our method preserves structural integrity and visual fidelity, showcasing its effectiveness in high-resolution video frame interpolation.

\subsection{Limitations and Future Work.}

\noindent\textbf{Limitations.} Although FC-VFI achieves lower inference times than existing diffusion-based methods, its Temporal Reference mechanism, which leverages boundary frames \(\mathbf{z}_s, \mathbf{z}_e\) to enhance intermediate frame quality, introduces additional computational overhead in the temporal dimension. The Matching-Lines Condition, limited by the video latent's multi-frame decoding nature, cannot explicitly control intermediate frames \(\{\mathbf{z}^n\}\) like FCVG \cite{zhu2025generative}. Instead, it enhances boundary frames' semantic information \[\mathbf{z}_s' = \mathbf{z}_s + \mathbf{c}_{\text{s}}, \mathbf{z}_e' = \mathbf{z}_e + \mathbf{c}_{\text{e}}\] to indirectly improve DiT's feature capture, constraining control flexibility. Currently, FC-VFI is validated only on HunyuanVideo-I2V, lacking evaluation on other models like CogVideoX \cite{yang2024cogvideox} or WAN \cite{wan2025}. Performance in complex dynamic scenes (e.g., water, smoke) is limited by training data and base model capabilities.

\noindent\textbf{Future Work.} We aim to enhance inference efficiency by reducing the current 10-step inference to 5 steps or single-step via model distillation and attention parallelization, narrowing the efficiency gap with optical flow methods at high resolutions (e.g., 2560$\times$1440). We plan to validate the Temporal Fidelity Modulation Reference (TFMR) framework on diverse architectures like WAN \cite{wan2025}, exploring its generalization. Additionally, extending training data and model capacity will enable 4K resolution and high-rate interpolation (16$\times$ or 32$\times$) for higher-fidelity VFI.

\clearpage
\end{document}

%% file: tables/quantitative_results.tex
\begin{table*}[t]
    \centering
    \setlength{\tabcolsep}{1.2mm}
        \resizebox{\linewidth}{!}{
    \begin{tabular}{lp{1.5cm}<{\centering}p{1.5cm}<{\centering}p{1.5cm}<{\centering}p{1.5cm}<{\centering}p{1.5cm}<{\centering}|p{1.5cm}<{\centering}p{1.5cm}<{\centering}p{1.5cm}<{\centering}p{1.5cm}<{\centering}p{1.5cm}<{\centering}} 
        \toprule
        \textbf{Methods} & \multicolumn{5}{c|}{$4\times$} & \multicolumn{5}{c}{$8\times$} \\
        \midrule
        \textbf{Optical-flow-based methods}& PSNR $\uparrow$ & SSIM $\uparrow$ & FID $\downarrow$ & FVD $\downarrow$ & LPIPS $\downarrow$  & PSNR $\uparrow$ & SSIM $\uparrow$ & FID $\downarrow$ & FVD $\downarrow$ & LPIPS $\downarrow$  \\
        GIMM-VFI~\cite{guo2024generalizable}                  & 29.05 & 0.901 & 16.22 & 125.42 & 0.061 & 29.49 & 0.907 & 14.75 & 192.36  & 0.048 \\
        Ours ($2560 \times 1440$)    & 30.25 & 0.915 & 15.73 & 130.65 & 0.054 & 30.16 & 0.912 & 15.50 & 194.19  & 0.046 \\
        \midrule
        \textbf{Diffusion-based methods}& PSNR $\uparrow$ & SSIM $\uparrow$ & FID $\downarrow$ & FVD $\downarrow$ & LPIPS $\downarrow$  & PSNR $\uparrow$ & SSIM $\uparrow$ & FID $\downarrow$ & FVD $\downarrow$ & LPIPS $\downarrow$  \\
        GI~\cite{wang2024generative}                           & 20.96 & 0.847 & 37.58 & 1310.80 & 0.119  & 21.05 & 0.694 & 39.24 & 940.72 & 0.128 \\
        ViBiDSampler~\cite{yang2024vibidsampler}                 & 23.48 & 0.764 & 31.92 & 1375.15 & 0.107  & 20.99 & 0.699 & 36.74 & 978.68 & 0.125 \\
        FCVG~\cite{zhu2025generative}                         & 26.70 & 0.830 & 20.12 & 330.04  & 0.055  & 25.80 & 0.811 & 21.79 & 251.10 & 0.059 \\
        \textbf{Ours ($1024 \times 576$)}   & \textbf{31.09} & \textbf{0.927} & \textbf{14.15} & \textbf{120.13} & \textbf{0.042} 
                                            & \textbf{31.21} & \textbf{0.917} & \textbf{14.03} & \textbf{187.10} & \textbf{0.041} \\
        \bottomrule
    \end{tabular}}
    \vspace{-0.1in}
    \caption{Quantitative comparison with optical-flow-based methods (evaluated at $2560 \times 1440$) and diffusion-based methods (evaluated at $1024 \times 576$) under $4\times$ and $8\times$ interpolation settings.}
    \label{tab_quantitative_results}
\end{table*}

%% file: tables/ablation.tex
\begin{table*}[!htbp]
    \centering
    \setlength{\tabcolsep}{1mm}
    \resizebox{\linewidth}{!}{
    \begin{tabular}{lp{1.3cm}<{\centering}p{1.3cm}<{\centering}p{1.3cm}<{\centering}p{1.3cm}<{\centering}p{1.3cm}<{\centering}|p{1.3cm}<{\centering}p{1.3cm}<{\centering}p{1.3cm}<{\centering}p{1.3cm}<{\centering}p{1.3cm}<{\centering}} 
        \toprule
        \textbf{Methods}& \multicolumn{5}{c|}{$4\times$} & \multicolumn{5}{c}{$8\times$} \\
        \hline
        \textit{Different Variants} & PSNR $\uparrow$ & SSIM $\uparrow$ & FID $\downarrow$ & FVD $\downarrow$ & LPIPS $\downarrow$ & PSNR $\uparrow$ & SSIM $\uparrow$ & FID $\downarrow$ & FVD $\downarrow$ & LPIPS $\downarrow$ \\
        \hline
        Channel Reference                                        & 26.07                  & 0.868                  & 68.16                 & 293.83                & 0.1067              & 24.73                     & 0.839                 & 70.19                 & 501.84                 & 0.1359          \\
        \hline
        Temporal Reference (our baseline)                        & 29.83                  & 0.921                  & 24.79                 & 185.67                & 0.0672              & 27.34                     & 0.893                 & 32.33                 & 407.10                 & 0.1021          \\
        + Fidelity Modulation (Sec.~\ref{sec:2})                 & 30.65                  & 0.925                  & \textbf{14.78}        & 178.82                & 0.0552              & \underline{28.69}         & 0.906                 & \underline{20.42}     & 326.37                 & \textbf{0.0832}         \\
        + $\mathcal{L}_{\text{temp}}$  (Sec.~\ref{sec:3})        & \underline{30.75}      & \underline{0.926}      & \underline{15.48}     & \underline{165.87}    & \textbf{0.0548}     & 28.67                     & \underline{0.907}     & \textbf{19.96}        & \underline{312.64}        & 0.0845                  \\
        + Matching Lines Condition  (Sec.~\ref{sec:4})           & \textbf{30.89}         & \textbf{0.928}         & 17.19                 & \textbf{153.34}       & \underline{0.0550}  & \textbf{29.13}            & \textbf{0.912}        & 21.44                 & \textbf{302.04}     & \textbf{0.0832}         \\
        \bottomrule
    \end{tabular}}
    \vspace{-0.1in}
    \caption{Ablation study demonstrating the impact of temporal reference (baseline), fidelity modulation, temporal difference loss $\mathcal{L}_{\text{temp}}$, and matching lines condition on interpolation quality. A channel reference (the variant using only channel-wise concatenation) is also included for comparison. Metrics are reported for $4\times$ and $8\times$ settings.}
    \label{tab_ablation}
    \vspace{-0.1in}
\end{table*}

%% file: tables/computation_efficiency.tex
\begin{table}[!htbp]
    \setlength{\tabcolsep}{1.5mm}
    \resizebox{\linewidth}{!}{
    \begin{tabular}{lcccc}
    \toprule
    \textbf{Method} & NFE & \makecell{Time (s) ($4\times$)} & \makecell{Time (s) ($8\times$)} & Resolution \\
    \midrule
    GI~\cite{wang2024generative}              & 300       & 606       & 606       & 1024 $\times$ 576 \\
    ViBiDSampler~\cite{yang2024vibidsampler}    & 50        & \underline{23}        & 38        & 1024 $\times$ 576 \\
    FCVG~\cite{zhu2025generative}           & 50        & 89        & 145       & 1024 $\times$ 576 \\
    \textbf{Ours}   & \textbf{10}       & \textbf{16}        & \textbf{22}        & 1024 $\times$ 576 \\
    \textbf{Ours}   & \textbf{10}       & 27                 & \underline{37}                  & \textbf{1280 $\times$ 720} \\
    \bottomrule
    \end{tabular}}
\vspace{-0.1in}
    \caption{Our method achieves significantly faster inference compared to other diffusion-based approaches.}
    \vspace{-0.1in}
    \label{tab_computation_efficiency}
    \end{table}

%% file: refs.bib
@inproceedings{wang2024framer,
  title={Framer: Interactive frame interpolation},
  author={Wang, Wen and Wang, Qiuyu and Zheng, Kecheng and Ouyang, Hao and Chen, Zhekai and Gong, Biao and Chen, Hao and Shen, Yujun and Shen, Chunhua},
  booktitle={ICLR},
  year={2025}
}

@inproceedings{wang2024generative,
  title={Generative inbetweening: Adapting image-to-video models for keyframe interpolation},
  author={Wang, Xiaojuan and Zhou, Boyang and Curless, Brian and Kemelmacher-Shlizerman, Ira and Holynski, Aleksander and Seitz, Steven M},
  booktitle={ICLR},
  year={2025}
}

@inproceedings{yang2024vibidsampler,
  title={Vibidsampler: Enhancing video interpolation using bidirectional diffusion sampler},
  author={Yang, Serin and Kwon, Taesung and Ye, Jong Chul},
  booktitle={ICLR},
  year={2025}
}

@inproceedings{zhu2025generative,
  title={Generative inbetweening through frame-wise conditions-driven video generation},
  author={Zhu, Tianyi and Ren, Dongwei and Wang, Qilong and Wu, Xiaohe and Zuo, Wangmeng},
  booktitle={CVPR},
  year={2025}
}

@inproceedings{feng2024explorative,
  title={Explorative inbetweening of time and space},
  author={Feng, Haiwen and Ding, Zheng and Xia, Zhihao and Niklaus, Simon and Abrevaya, Victoria and Black, Michael J and Zhang, Xuaner},
  booktitle={ECCV},
  year={2024}
}

@inproceedings{chen2025repurposing,
  title={Repurposing pre-trained video diffusion models for event-based video interpolation},
  author={Chen, Jingxi and Feng, Brandon Y and Cai, Haoming and Wang, Tianfu and Burner, Levi and Yuan, Dehao and Fermuller, Cornelia and Metzler, Christopher A and Aloimonos, Yiannis},
  booktitle={CVPR},
  year={2025}
}

@article{peng2024controlnext,
  title={Controlnext: Powerful and efficient control for image and video generation},
  author={Peng, Bohao and Wang, Jian and Zhang, Yuechen and Li, Wenbo and Yang, Ming-Chang and Jia, Jiaya},
  journal={arXiv preprint arXiv:2408.06070},
  year={2024}
}

@article{guo2024generalizable,
  title={Generalizable implicit motion modeling for video frame interpolation},
  author={Guo, Zujin and Li, Wei and Loy, Chen Change},
  journal={NeurIPS},
  year={2024}
}

@inproceedings{seo2025bim,
  title={BiM-VFI: Bidirectional Motion Field-Guided Frame Interpolation for Video with Non-uniform Motions},
  author={Seo, Wonyong and Oh, Jihyong and Kim, Munchurl},
  booktitle={CVPR},
  year={2025}
}

@inproceedings{danier2024ldmvfi,
  title={Ldmvfi: Video frame interpolation with latent diffusion models},
  author={Danier, Duolikun and Zhang, Fan and Bull, David},
  booktitle={AAAI},
  year={2024}
}

@article{blattmann2023stable,
  title={Stable video diffusion: Scaling latent video diffusion models to large datasets},
  author={Blattmann, Andreas and Dockhorn, Tim and Kulal, Sumith and Letts, Adam and others},
  journal={arXiv},
  year={2023}
}

@inproceedings{zhang2023adding,
  title={Adding conditional control to text-to-image diffusion models},
  author={Zhang, Lvmin and Rao, Anyi and Agrawala, Maneesh},
  booktitle={ICCV},
  year={2023}
}

@article{zhang2025motion,
  title={Motion-aware generative frame interpolation},
  author={Zhang, Guozhen and Zhu, Yuhan and Cui, Yutao and Zhao, Xiaotong and Ma, Kai and Wang, Limin},
  journal={arXiv preprint arXiv:2501.03699},
  year={2025}
}

@inproceedings{huang2022real,
  title={Real-time intermediate flow estimation for video frame interpolation},
  author={Huang, Zhewei and Zhang, Tianyuan and Heng, Wen and Shi, Boxin and Zhou, Shuchang},
  booktitle={ECCV},
  year={2022}
}

@article{kong2024hunyuanvideo,
  title={Hunyuanvideo: A systematic framework for large video generative models},
  author={Kong, Weijie and Tian, Qi and Zhang, Zijian and Min, Rox and Dai, Zuozhuo and Zhou, Jin and Xiong, Jiangfeng and Li, Xin and Wu, Bo and Zhang, Jianwei and others},
  journal={arXiv},
  year={2024}
}

@inproceedings{pautrat2023gluestick,
  title={Gluestick: Robust image matching by sticking points and lines together},
  author={Pautrat, R{\'e}mi and Su{\'a}rez, Iago and Yu, Yifan and Pollefeys, Marc and Larsson, Viktor},
  booktitle={CVPR},
  year={2023}
}

@inproceedings{hu2022lora,
  title={Lora: Low-rank adaptation of large language models.},
  author={Hu, Edward J and Shen, Yelong and Wallis, Phillip and Allen-Zhu, Zeyuan and Li, Yuanzhi and Wang, Shean and Wang, Lu and Chen, Weizhu and others},
  booktitle={ICLR},
  year={2022}
}

@inproceedings{peebles2023scalable,
  title={Scalable diffusion models with transformers},
  author={Peebles, William and Xie, Saining},
  booktitle={ICCV},
  year={2023}
}

@article{wan2025,
      title={Wan: Open and Advanced Large-Scale Video Generative Models}, 
      author={Team Wan and Ang Wang and Baole Ai and Bin Wen and Chaojie Mao and Chen-Wei Xie and Di Chen and Feiwu Yu and Haiming Zhao and Jianxiao Yang and Jianyuan Zeng and Jiayu Wang and Jingfeng Zhang and Jingren Zhou and Jinkai Wang and Jixuan Chen and Kai Zhu and Kang Zhao and Keyu Yan and Lianghua Huang and Mengyang Feng and Ningyi Zhang and Pandeng Li and Pingyu Wu and Ruihang Chu and Ruili Feng and Shiwei Zhang and Siyang Sun and Tao Fang and Tianxing Wang and Tianyi Gui and Tingyu Weng and Tong Shen and Wei Lin and Wei Wang and Wei Wang and Wenmeng Zhou and Wente Wang and Wenting Shen and Wenyuan Yu and Xianzhong Shi and Xiaoming Huang and Xin Xu and Yan Kou and Yangyu Lv and Yifei Li and Yijing Liu and Yiming Wang and Yingya Zhang and Yitong Huang and Yong Li and You Wu and Yu Liu and Yulin Pan and Yun Zheng and Yuntao Hong and Yupeng Shi and Yutong Feng and Zeyinzi Jiang and Zhen Han and Zhi-Fan Wu and Ziyu Liu},
      journal = {arXiv},
      year={2025}
}

@inproceedings{yang2024cogvideox,
  title={CogVideoX: Text-to-Video Diffusion Models with An Expert Transformer},
  author={Yang, Zhuoyi and Teng, Jiayan and Zheng, Wendi and Ding, Ming and Huang, Shiyu and Xu, Jiazheng and Yang, Yuanming and Hong, Wenyi and Zhang, Xiaohan and Feng, Guanyu and others},
  booktitle={ICLR},
  year={2025}
}

@article{xu2023conditional,
  title={Conditional temporal variational autoencoder for action video prediction},
  author={Xu, Xiaogang and Wang, Yi and Wang, Liwei and Yu, Bei and Jia, Jiaya},
  journal={IJCV},
  year={2023}
}

@inproceedings{pan2025boosting,
  title={Boosting Diffusion-Based Text Image Super-Resolution Model Towards Generalized Real-World Scenarios},
  author={Pan, Chenglu and Xu, Xiaogang and Ding, Ganggui and Zhang, Yunke and Li, Wenbo and Xu, Jiarong and Wu, Qingbiao},
  booktitle={ICCVW},
  year={2025}
}

@inproceedings{xu2022mtformer,
  title={Mtformer: Multi-task learning via transformer and cross-task reasoning},
  author={Xu, Xiaogang and Zhao, Hengshuang and Vineet, Vibhav and Lim, Ser-Nam and Torralba, Antonio},
  booktitle={ECCV},
  year={2022}
}

@article{lipman2022flow,
  title={Flow matching for generative modeling},
  author={Lipman, Yaron and Chen, Ricky TQ and Ben-Hamu, Heli and Nickel, Maximilian and Le, Matt},
  journal={arXiv preprint arXiv:2210.02747},
  year={2022}
}

@inproceedings{xu2022hierarchical,
  title={Hierarchical image generation via transformer-based sequential patch selection},
  author={Xu, Xiaogang and Xu, Ning},
  booktitle={AAAI},
  year={2022}
}

@inproceedings{ding2024freecustom,
  title={FreeCustom: Tuning-Free Customized Image Generation for Multi-Concept Composition}, 
  author={Ganggui Ding and Canyu Zhao and Wen Wang and Zhen Yang and Zide Liu and Hao Chen and Chunhua Shen},
  booktitle={CVPR},
  year={2024}
}

@article{ma2021bvi,
  title={BVI-DVC: A training database for deep video compression},
  author={Ma, Di and Zhang, Fan and Bull, David R},
  journal={IEEE Transactions on Multimedia},
  year={2021}
}

@inproceedings{nah2019ntire,
  title={Ntire 2019 challenge on video deblurring and super-resolution: Dataset and study},
  author={Nah, Seungjun and Baik, Sungyong and Hong, Seokil and Moon, Gyeongsik and Son, Sanghyun and Timofte, Radu and Mu Lee, Kyoung},
  booktitle={CVPRW},
  year={2019}
}

@inproceedings{shen2020blurry,
  title={Blurry video frame interpolation},
  author={Shen, Wang and Bao, Wenbo and Zhai, Guangtao and Chen, Li and Min, Xiongkuo and Gao, Zhiyong},
  booktitle={CVPR},
  year={2020}
}

@article{unterthiner2018towards,
  title={Towards accurate generative models of video: A new metric \& challenges},
  author={Unterthiner, Thomas and Van Steenkiste, Sjoerd and Kurach, Karol and Marinier, Raphael and Michalski, Marcin and Gelly, Sylvain},
  journal={arXiv},
  year={2018}
}

@article{heusel2017gans,
  title={Gans trained by a two time-scale update rule converge to a local nash equilibrium},
  author={Heusel, Martin and Ramsauer, Hubert and Unterthiner, Thomas and Nessler, Bernhard and Hochreiter, Sepp},
  journal={Advances in neural information processing systems},
  year={2017}
}

@inproceedings{zhang2018unreasonable,
  title={The unreasonable effectiveness of deep features as a perceptual metric},
  author={Zhang, Richard and Isola, Phillip and Efros, Alexei A and Shechtman, Eli and Wang, Oliver},
  booktitle={CVPR},
  year={2018}
}

@inproceedings{sim2021xvfi,
  title={Xvfi: extreme video frame interpolation},
  author={Sim, Hyeonjun and Oh, Jihyong and Kim, Munchurl},
  booktitle={ICCV},
  year={2021}
}

@article{pont20172017,
  title={The 2017 davis challenge on video object segmentation},
  author={Pont-Tuset, Jordi and Perazzi, Federico and Caelles, Sergi and Arbel{\'a}ez, Pablo and Sorkine-Hornung, Alex and Van Gool, Luc},
  journal={arXiv preprint arXiv:1704.00675},
  year={2017}
}

@article{yang2023object,
  title={Object-aware inversion and reassembly for image editing},
  author={Yang, Zhen and Ding, Ganggui and Wang, Wen and Chen, Hao and Zhuang, Bohan and Shen, Chunhua},
  journal={arXiv preprint arXiv:2310.12149},
  year={2023}
}

@article{wang2004image,
  title={Image quality assessment: from error visibility to structural similarity},
  author={Wang, Zhou and Bovik, Alan C and Sheikh, Hamid R and Simoncelli, Eero P},
  journal={IEEE transactions on image processing},
  volume={13},
  number={4},
  pages={600--612},
  year={2004},
  publisher={IEEE}
}

@article{kingma2013auto,
  title={Auto-encoding variational bayes},
  author={Kingma, Diederik P and Welling, Max},
  journal={arXiv preprint arXiv:1312.6114},
  year={2013}
}
